\documentclass{article} 
\usepackage{iclr2023_conference,times}

\usepackage{amsmath,amsfonts,bm}

\def\eqref#1{equation~\ref{#1}}

\def\1{\bm{1}}

\def\vp{{\bm{p}}}

\def\vx{{\bm{x}}}

\def\mP{{\bm{P}}}
\def\mQ{{\bm{Q}}}

\DeclareMathAlphabet{\mathsfit}{\encodingdefault}{\sfdefault}{m}{sl}
\SetMathAlphabet{\mathsfit}{bold}{\encodingdefault}{\sfdefault}{bx}{n}

\def\gG{{\mathcal{G}}}

\usepackage{hyperref}
\usepackage{url}
\usepackage{comment}
\usepackage{tabularx}

\title{Diffsurv: Differentiable sorting for censored time-to-event data.}

\author{
Andre Vauvelle\textsuperscript{1}, Benjamin Wild\textsuperscript{2}, Aylin Cakiroglu\textsuperscript{3}, Roland Eils\textsuperscript{2}, Spiros Denaxas\textsuperscript{1} \\
University College London\textsuperscript{1}, Berlin Institute of Health\textsuperscript{2},  BenevolentAI\textsuperscript{3}, \\
\texttt{\{rmhivau,s.denaxas\}@ucl.ac.uk}, \texttt{aylin.cakiroglu@benevolent.ai}, \\ \texttt{\{benjamin.wild, roland.eils\}@bih-charite.de}
}

\usepackage{booktabs}
\usepackage{amsmath}
\usepackage{amssymb}
\usepackage{natbib}
\usepackage{xcolor}
\usepackage{multirow}
\usepackage{bbm}
\usepackage{lipsum}
\usepackage[position=top]{subfig}
\usepackage{lineno}
\usepackage{graphicx}
\usepackage{wrapfig}
\usepackage{tikz}

\def\reals{\mathbb{R}}

\iclrfinalcopy 
\begin{document}


\maketitle

\begin{abstract}
Survival analysis is a crucial semi-supervised task in machine learning with numerous real-world applications, particularly in healthcare. Currently, the most common approach to survival analysis is based on Cox's partial likelihood, which can be interpreted as a ranking model optimized on a lower bound of the concordance index. This relation between ranking models and Cox’s partial likelihood considers only pairwise comparisons. Recent work has developed differentiable sorting methods which relax this pairwise independence assumption, enabling the ranking of sets of samples. However, current differentiable sorting methods can not account for censoring, a key factor in many real-world datasets. To address this limitation, we propose a novel method called \emph{Diffsurv}. We extend differentiable sorting methods to handle censored tasks by predicting matrices of possible permutations that take into account the label uncertainty introduced by censored samples. We contrast this approach with methods derived from partial likelihood and ranking losses. Our experiments show that Diffsurv outperforms established baselines in various simulated and real-world risk prediction scenarios. Additionally, we demonstrate the benefits of the algorithmic supervision enabled by Diffsurv by presenting a novel method for top-k risk prediction that outperforms current methods.
\end{abstract}

\section{Introduction and Background}

Survival analysis is an important task in numerous machine learning applications, particularly in the healthcare domain. The goal of survival analysis is to predict the time until the occurrence of an event of interest, such as death or disease onset, based on a set of covariates. In clinical studies, these covariates typically include demographic variables such as sex and age, but may also encompass more complex data modalities such as temporal streams or medical images.

However, event times may not be observed due to censoring, especially in observational datasets where many patients may not have experienced the event at the time of data collection. Ignoring censoring can lead to biased predictions towards the censoring event instead of the event of interest. For example, if the end of the study can be determined from the observed covariates, especially if age is included, the predicted event times will be skewed towards the censoring event time instead of the actual event of interest \citet{kvamme_brier_2019}.

The Cox Proportional Hazards (PH) model is widely used for handling censored data in survival analysis \citep{cox_regression_1972}. The model optimizes a partial likelihood function over ranked data, considering only the order of events, not their exact time of occurrence. As such, Cox's partial likelihood serves as a ranking loss, learning from the order of patients based on their hazard of experiencing an event, not their exact survival time. \citet{raykar_ranking_2007} showed that Cox PH and ranking models can be directly equated, with both providing lower bounds to the concordance index, the primary evaluation metric used in survival analysis. A key step in relating the two models assumes only pairwise comparisons or risk sets of size 2. \citet{goldstein_asymptotic_1992} show that sub-sampling risk sets produce consistent parameter estimators but that greater risk sets provide more efficient estimators. Cox's partial likelihood and ranking losses underpin current survival analysis methods in deep learning, including DeepSurv \citet{katzman_deepsurv_2018} and DeepHit \citet{lee_deephit_2018}.


We present an alternative method that leverages recent advancements in continuous relaxations of sorting operations, enabling end-to-end training of neural networks with ordering supervision \citep{grover_stochastic_2019, blondel_fast_2020, petersen_differentiable_2021}. This involves incorporating a sorting algorithm into the network architecture, where the order of the samples is known, but their exact values are unsupervised. Here, we introduce \emph{Diffsurv}, an extension of differentiable sorting methods that enables end-to-end training of survival models with censored data.

Briefly, our contributions are summarised:
\begin{itemize}
    \item Our primary contribution is the extension differentiable sorting methods to account for censoring by introducing the concept of possible permutation matrices. 
    \item We empirically demonstrate that our new differentiable sorting method improve risk ranking performance across multiple simulated and real-worlds censored datasets.
    \item We demonstrate that differentiable sorting of censored data enables the development of new methods with practical applications, using the example of end-to-end learning for top-k risk stratification.
\end{itemize}

\section{Methods}

\begin{figure}
    \centering
    \includegraphics[width=\linewidth]{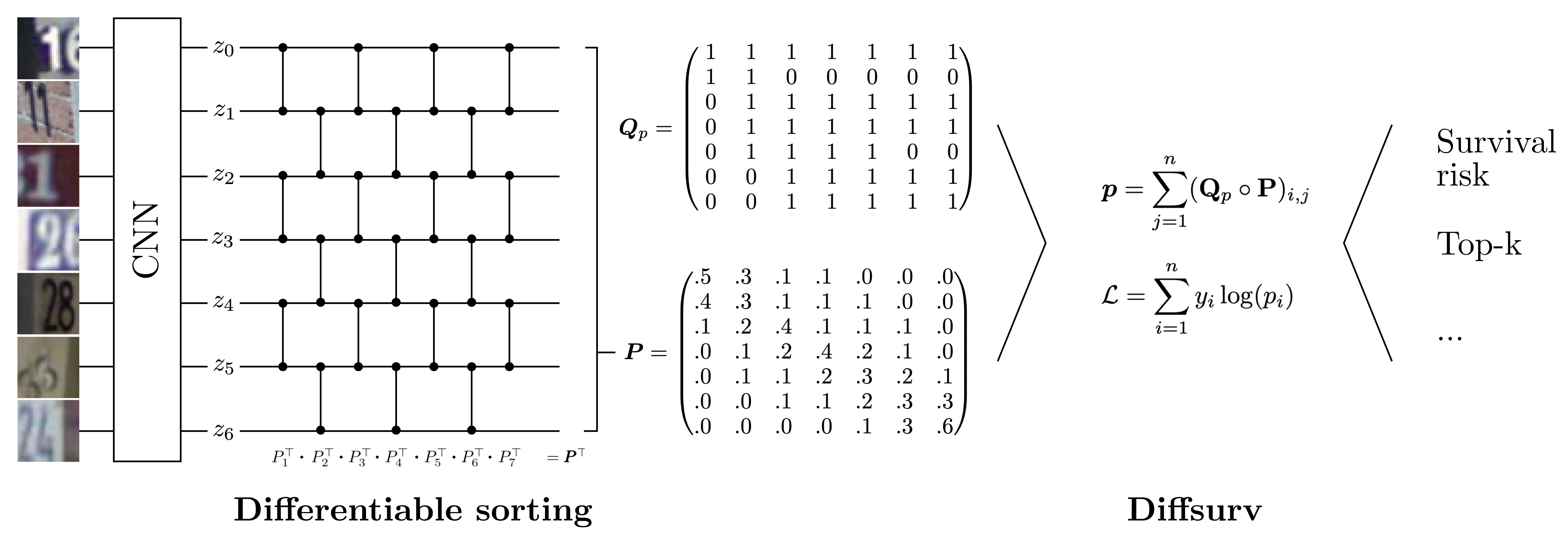}
    \caption{Differentiable Sorting for Censored Time-to-Event Data. Inputs, in this case, SVHN images, are transformed into scalar risk values, $z_i$, through a neural network. A differentiable permutation matrix, $\mP$, is computed using sorting networks. The model can be optimized for downstream tasks, such as risk stratification and top-k highest risk prediction, by using the matrix $\mQ_p$ of possible permutations based on the observed events and censoring.}
    \label{fig:abstract}
\end{figure}
A dataset with censored event times is summarized as $\mathcal{D}= \{t_i, \vx_i, \delta_i\}_{i=1}^N$. For a patient $i$, $t_i$ is the observed minimum of the unobserved true survival time $t_i^*$ and the censoring time $c_i^*$, $\delta_i$ is the event indicator that is 1 if an event is observed ($t_i^* \leq c_i^*$) or 0 if the data is censored ($t_i^* > c_i^*$). Covariates are $\vx_i \in \reals^{d}$ representing a 1-dimensional vector of size $d$ or larger dimensional tensors such as image data. $N$ is the total number of patients. As previously mentioned, it is common to subsample total possible risk set, we use $n$ to represent the subsampled risk set size. 

In order to train models based on ordering information using differentiable sorting algorithms \citet{petersen_learning_2022}, we can minimize the cross-entropy between the ground truth orders represented by true permutation matrix $\mQ$ and a doubly-stochastic predicted permutation matrix $\mP$. 
This makes it possible to interpret each element $P_{ij}$ of the predicted permutation matrix as the predicted probability of permuting from a randomly assigned rank $i$ to a true rank $j$.  

There are multiple methods of relaxing sorting algorithms to produce $\mP$, we will follow \citet{petersen_differentiable_2021} by using differentiable sorting networks. Sorting networks are a family of sorting algorithms that consist of two basic components: wires and conditional swaps.  Wires carry values to be compared at conditional swaps, if one value is bigger than the other, the values carried forward are swapped around. For a random sample of patients to be ordered, each layer of the sorting network can be considered an independent permutation matrix $\mP_l$ with elements given by
\begin{equation}
    P_{l,ii} = P_{l,jj} = \sigma(z_j - z_i) \text{   and   } 
    P_{l,ij} = P_{l,ji} = 1 - \sigma(z_j - z_i).
\end{equation}


These elements represent conditional swaps between two patient risk values $(z_i,z_j)$ and use a differentiable relaxation of the step function such as the logistic-sigmoid, where $\sigma: x \rightarrow \frac{1}{1+e^{-\beta x}}$. The inverse temperature parameter $\beta>0$ is introduced so when $\beta \rightarrow \infty$ the functions tend to the exact $\min$ and $\max$ functions. The indices being compared are determined by the sorting network and the final predicted probability matrix is the product of each layer of sorting operations, $\mP = \left(\prod_{l=1}^n \mP_l^\intercal\right)^\intercal$. For the base case, $n=2$, Diffsurv is equivalent to the pairwise ranking loss and Cox partial likelihood. Further details on the relations between Diffsurv and baselines is in Appendix~\ref{apd:relations}.

The introduction of censored patients means we no longer have access to a ground truth permutation matrix $\mathbf{Q}$. We cannot determine the exact rank of patients who are censored before another who experienced an event. To address this challenge, we propose a novel extension of differentiable sorting to censored data. Our approach considers the set of \emph{possible permutations} for each patient, taking into account uncertainty about the true ranking. In Figure~\ref{fig:possible-permutations}, we show an example of observed and censored events and the resulting set of possible permutations that can be represented as a permutation matrix $\mQ_p$.

For a right-censored sample $i$, we only know that the rank must be lower than the rank of all other samples with an event time lower than the censoring time of $t_i$, i.e. they must be ranked after prior events. For another sample $j$ with an event at $t_j$, we know that the rank must be lower than other samples with an event time lower than $t_j$, and higher than the rank of other samples either with an event time higher than $t_j$ or with a censoring time higher than $t_j$. We do not know how the rank of $j$ compares to samples with censoring time lower than $t_j$. If it is possible for patient i to permute to rank j, then $Q_{pij}=1$, otherwise $Q_{pij}=0$. 

Given the possible permutation matrix $\mathbf{Q}_p$ and the predicted permutation matrix $\mathbf{P}$, the vector of probabilities $\vp$ of a patient being ranked within the set of possible permutations can be computed. Although the ground truth ranks are unknown, the range of possible ranks is known, and the model can be optimized to maximize the sum of the predicted permutation probabilities for the possible ranks of each sample. Noted here as the column-sum of the element-wise product $\circ$, between $\mathbf{Q}_p$ and $\mathbf{P}$.
\begin{equation}
    \vp = \sum_{j=1}^n (\mathbf{Q}_p \circ \mathbf{P})_{i,j}.
\end{equation}
The cross-entropy loss can then be easily applied
\begin{equation}
    \mathcal{L} = \sum_{i=1}^n y_{i} \log (p_i)
\end{equation}
where $y_i$ is the true label of the set of possible ranks. 


Finally, we demonstrate how the algorithmic supervision of sorting algorithms enables the development of novel methods in survival analysis, using the example of top-k risk prediction. In practical settings, it is often not necessary to rank all samples correctly. Rather, it is essential to identify the samples with the highest risk, such as by a healthcare provider, to prioritize care and interventions. With Diffsurv, top-k risk prediction is straightforward to implement by optimizing possible permutations within the top-k ranks, whereby $\mQ_p$ is adjusted such that only the top-k patient's possible permutations are set to 1. 
 

\section{Experiments}

We evaluate the performance of Diffsurv on censored survival data across semi-synthetic and real-world datasets. In each experiment, we train a neural network using Diffsurv and Cox's partial likelihood loss, then compare their respective results. Cox's partial likelihood and the closely related ranking loss are used in popular baselines; Deepsurv \citep{katzman_deepsurv_2018}, Cox-MLP \citep{kvamme_time--event_2019} and DeepHit  \citet{lee_deephit_2018}. 

We present a new semi-synthetic dataset, \emph{survSVHN}, to evaluate survival models. Based on the Street View House Numbers (SVHN) dataset \cite{petersen_differentiable_2021}, we simulate survival times akin to survMNIST \cite{polsterl_survival_2019}. The increased complexity of SVHN offers a testbed which is better able to discern the performance differences between methods. Each house number parameterizes an exponential time function for survival times. Risks are calculated as the logarithm of house numbers, standardized and scaled for a mean survival time of 30. We introduce censoring by randomly selecting 30\% of house numbers and replacing true times with values sampled uniformly between $(0,t_i]$ (See Figure~\ref{fig:svnh-abstract}). Risk is predicted from the images with a convolutional neural network with the same hyperparameters as \cite{petersen_differentiable_2021}, with $z_i = f_\text{CONV}(\vx_i)$. 
We also evaluate on four real-world healthcare datasets from \citet{kvamme_time--event_2019}. Each dataset has a fairly small number of patients ($N\leq8,873$) and a flat vector of covariates as input. Further details in Appendix~\ref{apd:realworld}. For these datasets, a fully connected neural network is used to find the risk, $z_i = f_{\text{MLP}}(\vx_i)$. Further details on the training and evaluation procedures can be found in Appendix~\ref{apd:traineval}.

\begin{table}[t]
    \caption{Results for semi-synthetic and real-world datasets. Bold indicates significantly higher performance (t-test with a significance level of 0.01).}
    \label{tab:combined-results}
    \centering
    
    \subfloat[Semi-synthetic survSVHN Dataset Results. Mean (and standard deviation) over 5 trails with different seeds. Metric is C-index. \textsuperscript{\dag} When $n=2$ both methods are equivalent to the ranking loss.]{%
        \label{tab:svhn-results}
        \begin{tabularx}{\linewidth}{lXXXXX}
            \toprule
             Method & n=2\textsuperscript{\dag} & n=4 & n=8 & n=16 & n=32 \\
             \midrule
             Diffsurv & .918 (.003) & \textbf{.934} (.002) & \textbf{.940} (.001) & \textbf{.943} (.002) & \textbf{.941} (.002) \\
             Cox Partial Likelihood & .913 (.002) & .925 (.002) & .931 (.002) & .933 (.002) & .930 (.003) \\
        \bottomrule
        \end{tabularx}
    }
    
    \subfloat[Real-world datasets results. Mean (and standard deviation) over 5-folds measured in C-Index for the risk stratification task and proportion correctly predicted for the Top-k task.]{%
        \label{tab:real-world}
        \begin{tabularx}{\linewidth}{lXXXX}
            \toprule
            & FLCHAIN & NWTCO & SUPPORT & METABRIC \\
            \midrule
            Size & 6,524 & 4,028 & 8,873 & 1,904 \\
            Censored Proportion & 69.9\% & 85.8\% & 32.0\% & 42.1\% \\
            \midrule\midrule
            Risk Stratification \\
            \midrule
            Diffsurv & .787 (.012) & .691 (.018) & \textbf{.599} (.004) & .623 (.012) \\
            Cox Partial Likelihood & .787 (.016) & .690 (.014) & .584 (.004) & .615 (.018) \\
            \midrule
            \midrule
            Top 10\% prediction \\
            \midrule
            Diffsurv (Top-k)       &  \textbf{.952} (.014) &  .926 (.007) &  \textbf{.571} (.012) &  \textbf{.771} (.059) \\
            Cox Partial Likelihood &  .937 (.014) &  .919 (.026) &  .478 (.032) &  .561 (.039) \\
            \bottomrule
        \end{tabularx}
    }
    
\end{table}

The results presented in Table~\ref{tab:combined-results} demonstrates that Diffsurv achieves equal to or better performance on all datasets analyzed. Additionally, when Diffsurv is optimized for predicting the top 10\% of highest risk individuals, it outperforms Cox's partial likelihood on all four datasets. There is a significant improvement in the top 10\% highest-risk prediction when comparing models based on Cox's Partial Likelihood ($\mu=.825$, $\sigma=.005$) and Diffsurv optimized for Top-k prediction ($\mu=.944$, $\sigma=.008$) on the survSVHN dataset. 

\section{Conclusion}

Diffsurv represents a significant step in the field of survival analysis with censored data. Our experiments demonstrate the effectiveness of differentiable sorting methods in improving survival analysis predictions, particularly in censored datasets with Diffsurv matching or improving perfomance against Cox partial likelihood on all datasets. Additionally, Diffsurv has the potential to drive the development of new methods, such as the top-k risk stratification method presented in this work. It is noteworthy that while our method has shown promising results, further investigation is necessary to fully understand its potential and limitations. For instance, it would be valuable to examine the scalability of the method with larger real-world datasets and its capability to handle more complex censored scenarios. Further research could also investigate the integration of Diffsurv into clustering models. With its ability to handle censored data and its end-to-end training capability, Diffsurv presents a promising approach to survival analysis and holds great potential for enhancing risk prediction in real-world applications.

\bibliography{iclr2023_conference}
\bibliographystyle{iclr2023_conference}

\appendix
\section{Appendix}

\subsection{Cox's Partial Likelihood and ranking losses}

\label{apd:cox}

\citet{cox_regression_1972} introduced the most popular method for addressing censoring for survival analysis. The original work and many subsequent extensions seek to maximize the partial likelihood, the general form is described as

\begin{equation}
    \mathcal{L(\theta)} = \prod_{i: \delta = 1} \frac{f_\theta (\vx_i)}{\sum_{j:T_j>T_i} f_\theta(\vx_j)},
\end{equation}
where $f_\theta$ is the hazard function, a real valued score prediction function estimating the probability of an event at particular time, given input features $\vx_i$. The product is taken over the set of uncensored patients, while the denominator term considers only comparable pairs and includes censored patients with $T_j> T_i$.

The classic Cox proportional hazards model used $f_\theta = \text{exp}(\theta \cdot \vx_i)$. Multiple extensions of this basic loss relax the linear covariate interaction and proportional hazards assumptions by altering $f_\theta$. For example, \citep{katzman_deepsurv_2018} parameterized with a neural network $h_{\theta}$, such that $f_\theta=\text{exp}(h_\theta(\vx_i))$, to model non-linear interactions between covariates on the hazard. \citep{kvamme_time--event_2019} further show that this can be extended to non-proportional hazards by introducing temporal covariates, $f_\theta=\text{exp}(h_\theta(\vx_i, T_i))$. 

\citet{kvamme_time--event_2019} also make a few adjustments to the original partial likelihood loss. First, they consider that the risk set $\mathcal{R} = \{j: T_j>T_i \}$ is intractable for deep learning applications as it considers all comparable patients. Instead, it is possible take a fixed size sample risk set $\lvert \Tilde{\mathcal{R}} \rvert=n<N$ and further, it is reasonable to take a constant sample size of 1  and include the individual $i$ in the risk set (such that n=2). This leads to the simplified loss of the form

\begin{equation}
    \mathcal{L(\theta)} = \prod_{i: \delta = 1} \frac{f_\theta (\vx_i)}{f_\theta(\vx_i) + f_\theta(\vx_j)}, \text{  } j \in \mathcal{R}\setminus \{i\}.
\end{equation}

Further, we can take the mean log partial likelihood to be

\begin{equation}
    \text{loss} = \frac{1}{n_e} \sum_{i: \delta = 1} \log(1+\exp[h_\theta(\vx_j) - h_\theta(\vx_i)]), \text{  } j \in \mathcal{R}\setminus \{i\},
 \label{eq:kvamme-pl}
\end{equation}

where $n_e$ is the number of non-censored events. From this simplified form, it can be seen that the partial likelihood only considers the relative ordering or ranking of survival times. 

The concordance index or c-index \citet{harrell_evaluating_1982} is a commonly used as an evaluation for survival analysis methods and is a generalization of the Area Under the Receiver Operating Characteristic Curve (AUROC) that handles right-censored data.

\begin{equation}
 \text{c-index} = \frac{1}{n} \sum_{i: \delta = 1} \mathbbm{1}(f(\vx_i) < f(\vx_j)), j \in \mathcal{R} \setminus \{i\}. 
 \label{eq:cindex}
\end{equation}

\citet{raykar_ranking_2007} showed that the Cox's partial likelihood is approximately equivalent to maximizing the concordance index or C-index and that closer bounds can be found by maximizing the general ranking loss 

\begin{equation}
    \text{ranking-loss} = \frac{1}{\lvert{\mathcal{A}\rvert}}\sum_{(\vx_i, \vx_j) \in \mathcal{A}} \phi(f_\theta(\vx_i) - f_\theta(\vx_j)),
    \label{eq:ranking}
\end{equation}
where $\phi$ is a function that relaxes the non-differentiable $\mathbbm{1}$ of the C-index. We have also introduced $\mathcal{A}$ as the graph of acceptable pairs, where each node is a patient that can only be linked to another with an edge if we are sure that the first event occurs before the second. This is another way of describing the risk sets introduced earlier. From Equation~\ref{eq:kvamme-pl} is can be seen that $\phi: x \rightarrow -\log (1+\exp(-x)) = \log(\sigma(x))$. Here, we have shown that the simplifications to the partial likelihood made by \citet{kvamme_time--event_2019} are equivalent to using the log-sigmoid ranking loss.

The key difference between ranking and partial likelihood losses comes when considering the assumption that it is reasonable to take a constant sample size of 1 (one pair in the risk set) in the partial likelihood. This effectively introduces the assumption that each pair (i, j)  is independent of any other pair. However, this assumption seems puzzling given the inherent transitivity of ranking (if $i>j$ and $j>k$ then $i>k$). 


\subsection{Differentiable Sorting}

\begin{figure}
    \centering
    \subfloat[\label{fig:odd-even-network}]{%
        \resizebox{0.5\columnwidth}{!}{
            \begin{tikzpicture}
            \foreach \a in {1,...,8}
              \draw[thick] (0,\a) -- ++(9,0);
            \foreach \x in {1, ..., 8}
                \foreach \y in {2, ..., 7} {
                     \filldraw ({\x, \y}) circle (1.5pt);
                }
            \foreach \x in {1, 3, ..., 7}
                \foreach \y in {1, 8} {
                     \filldraw ({\x, \y}) circle (1.5pt);
                }
            \foreach \x in {1, 3, ..., 8}
                \foreach \y in {8, 6, ..., 2} {
                    \draw[thick] (\x,\y) -- (\x,\y-1);
                }
            \foreach \x in {2, 4, ..., 8}
                \foreach \y in {7, 5, ..., 3} {
                    \draw[thick] (\x,\y) -- (\x,\y-1);
                }
            \end{tikzpicture}
        }
    }
    \subfloat[\label{fig:bitonic-network}]{%
        \resizebox{0.5\columnwidth}{!}{
            \begin{tikzpicture}
            \foreach \a in {1,...,8}
              \draw[thick] (0,\a) -- ++(9,0);
            \foreach \x in {1, 3.66, 8}
                \foreach \y in {1, ..., 8} {
                     \filldraw ({\x, \y}) circle (1.5pt);
                }
            \foreach \x in {1, 3.66, 8}
                \foreach \y in {8, 6, ..., 2} {
                    \draw[thick] (\x,\y) -- (\x,\y-1);
                }
            \foreach \x in {2.33}
                \foreach \y in {1, 5} {
                     \filldraw ({\x, \y}) circle (1.5pt);
                     \filldraw ({\x, \y+3}) circle (1.5pt);
                }
            \foreach \x in {2.33}
                \foreach \y in {1, 5} {
                    \draw[thick] (\x,\y) -- (\x,\y+3);
                }
            \foreach \x in {2.66}
                \foreach \y in {2, 6} {
                     \filldraw ({\x, \y}) circle (1.5pt);
                     \filldraw ({\x, \y+1}) circle (1.5pt);
                }
            \foreach \x in {2.66}
                \foreach \y in {2, 6} {
                    \draw[thick] (\x,\y) -- (\x,\y+1);
                }
            \foreach \x in {0,...,3}
                \foreach \y in {1} {
                     \filldraw ({\x*0.33+4.66, \x+\y}) circle (1.5pt);
                }
            \foreach \x in {0,...,3}
                \foreach \y in {8} {
                     \filldraw ({\x*0.33+4.66, \y-\x}) circle (1.5pt);
                }
            \draw[thick] (4.66, 1) -- (4.66,8);
            \draw[thick] (1*0.33+4.66, 2) -- (1*0.33+4.66,7);
            \draw[thick] (2*0.33+4.66, 3) -- (2*0.33+4.66,6);
            \draw[thick] (3*0.33+4.66, 4) -- (3*0.33+4.66,5);

            \foreach \y in  {1,3} {
                 \filldraw ({6.44, \y}) circle (1.5pt);
                 \filldraw ({6.44, 4+\y}) circle (1.5pt);
                 \filldraw ({6.77, 1+\y}) circle (1.5pt);
                 \filldraw ({6.77, 5+\y}) circle (1.5pt);
             }

             \foreach \y in {1,5} {
                 \draw[thick] (6.44, \y) -- (6.44, \y+2);
                 \draw[thick] (6.77, \y+1) -- (6.77, \y+3);
             }
            \end{tikzpicture}
        }
    }
    \caption{Example sorting networks of size 8; (a) Odd-Even, (b) Bitonic.}
    \label{fig:sorting-networks}
\end{figure}
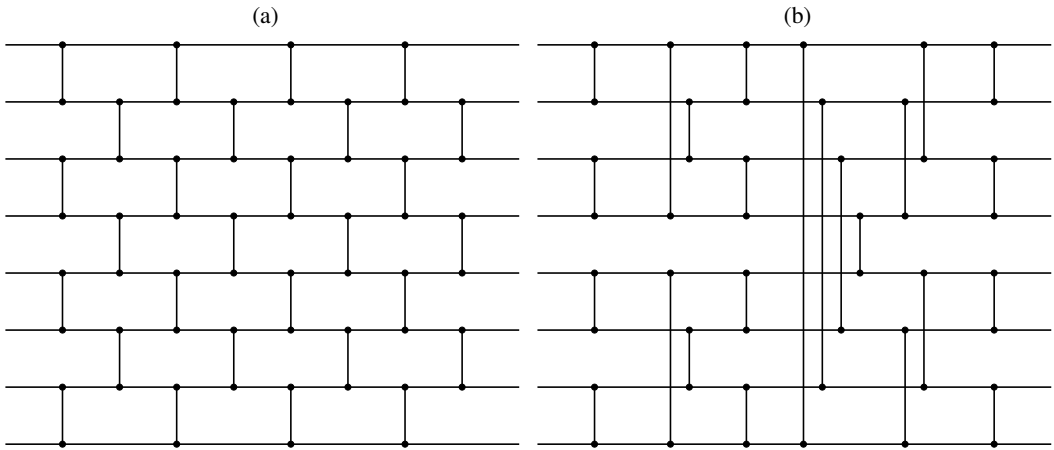

Differentiable sorting takes a different approach to the previously discussed partial likelihood loss and will need modification to account for censoring. In order train models based on ordering information, differences between predicted and true orderings are backpropagated through a relaxed sorting algorithm. The idea was first introduced by \citep{grover_stochastic_2019}, the key motivation being that sorting is a key step in many classical algorithms and machine learning methods (K-nearest neighbours), yet is non-differentiable. This means that direct supervision cannot be done on the outputs of algorithms that rely on sorting \citep{petersen_learning_2022}. 

Sorting algorithms require use of non-differentiable $\max$ and $\min$ operators. These are analogous to the non-differentiable indicator function that was discussed earlier in the c-index (Equation~\ref{eq:cindex}). Differentiable sorting methods similarly rely on approximating these operators with smooth alternatives.

\citet{petersen_differentiable_2021} propose combining traditional sorting networks and differentiable sorting functions. Sorting networks are a family of sorting algorithms that consist of two basic components: wires and conditional swaps. Wires carry values to be compared at conditional swaps, if one value is bigger than the other then the values carried forward are swapped around. This allows construction of provably guaranteed sorting networks.  Conditional swaps are exactly the min and max operators that ensure that with inputs $\{a,b\}$ and outputs $a^* \leq b^*$, $a^* = \min(a,b)$ and $b^* = \max(a,b)$. Note that as $a$ approaches $b$, the point at which it becomes larger is discontinuous and hence non-differentiable. Just as previously shown in the ranking loss, such operations can be made differentiable using the logistic relaxation
\begin{equation}
    \text{min}_\sigma(a,b) = a \cdot \sigma(b-a) + b \cdot \sigma(a-b) \text{   and   } \text{max}_\sigma(a,b) = a \cdot \sigma(a-b) + b \cdot \sigma(b-a).
\end{equation}
Note that if an inverse temperature parameter $\beta>0$ is introduced such that $\sigma: x \rightarrow \frac{1}{1+e^{-\beta x}}$, then as $\beta \rightarrow \infty$ the functions tend to the exact $\min$ and $\max$ functions. Other relaxation of the step function can also be considered, \citet{petersen_differentiable_2021} show that the Cauchy distribution preserves monotonicity which is desirable for optimization. Given this, we use the Cauchy distribution as our relaxation for all experiments, where $\sigma: x \rightarrow \frac{1}{\pi} \text{arctan}(\beta x) + \frac{1}{2}$.

There are multiple different types of sorting networks each with varying space complexity. The ability to implement networks with the divide-and-conquer paradigm allows for sorting networks that scale more efficiently than previous differentiable sorting methods. In particular, \citet{petersen_differentiable_2021} uses the odd-even and bitonic sorting networks. The latter allows construction of networks with size complexity $\mathcal{O}(nlog^2n)$ verses the $\mathcal{O}(n^2)$ in previous differentiable sorting methods. Examples for Odd-Even and Bitonic sorting networks with $n=8$ are shown in Figure~\ref{fig:sorting-networks}.

\subsubsection{Relation to ranking and partial likelihood}
\label{apd:relations}

Differentiable sorting has so far only been applied in the context of uncensored ranking, where we know the true rank for every instance in an input set. It is possible to directly relate Diffsurv with ranking losses. Expanding out the cross entropy loss out we find
\begin{equation}
    \mathcal{L} = \sum_c^n \left(\frac{1}{n} \sum_{i}^n q_{ci} \log (p_{ci}) \right),
\end{equation}
where $q_{ci}=1$ only when $i$ is the true rank otherwise $0$. Each $p_{ci}$ is always a function of the difference in pairs of inputs $x_i$ and $x_j$. This is complicated by the products of intermediate values $a_i$ introduced by the sorting network but denoted as

\begin{equation}
    p_{ci} = \prod^n_{(a_i, a_j)\in \mathcal{P}_l:  l=1}\sigma(f_\theta(a_i) - f_\theta(a_j))
\end{equation}
where $\mathcal{P}_l$ to denotes the set of comparisons to be made at each layer of the sorting network. With risk set of size 2, the loss returns to the same recognisable log-sigmoid ranking loss, and Cox negative log partial likelihood with risk set size 2.

\subsection{Non-proportional Hazards}

Our current implementation of Diffsurv is limited to proportional hazards, which may not fully capture the complexity of certain survival analysis problems, particularly when non-proportional hazards are present. In this context, we explore alternative approaches that address this limitation.

Previously, we briefly mentioned continuous-time extensions of partial likelihood to enable non-proportional hazards \citet{kvamme_time--event_2019}. This can be achieved by directly modeling temporal covariates as $f_\theta = \text{exp}(h_\theta(x_i, T_i))$.

Another class of methods focuses on discretizing the time-to-event variable and modeling the probability mass function (PMF) of event times. For instance, the DeepHit model \citet{lee_deephit_2018} employs a neural network architecture to learn the relationships between input features and discretized time-to-event outcomes. Time discretization facilitates modeling of non-proportional hazards but introduces two significant challenges: 1) sensitivity to the choice of time intervals, which can affect the model's accuracy and interpretability, and 2) increased computational complexity, as predictions must be made for each time interval. These models can be computationally expensive, especially for deep learning-based models like DeepHit, making them less suitable for high-dimensional and large-scale datasets, such as the imaging dataset used in this study.

Several future work proposals arise from these observations. First, differentiable sorting could explore the approach of directly modeling temporal covariates, resulting in a time-parameterized predicted permutation matrix. Second, extending Diffsurv to discrete time could be achieved by parameterizing a predicted permutation matrix for each time discretization. Finally, in models like DeepHit, the ranking loss term could be replaced with a Diffsurv loss, offering another promising direction for future research in survival analysis.

\subsection{Training and evaluation}

\label{apd:traineval}

During training, we use the Adam optimizer \citet{kingma_adam_2017} with a learning rate of $10^{-3}$, early stopping with patience of 10 epochs and a maximum of $10^5$ training steps. As in \citet{goldstein_asymptotic_1992} and \citet{kvamme_time--event_2019}, we ensure that each risk set contains a valid risk set by sampling controls for a given case. Each batch consists of a number of risk sets such that the input data has shape (batch size, risk set size, covariate shape). 

For the survSVHN task, the hazard function $h_\theta$ for both the Cox Partial Likelihood baseline and $f_\theta$ for Diffsurv, is a fixed sized convolutional neural network with a batch size of 100. Steepness is determined by risk set size $n$, $\beta=2n$ for Odd-even networks and $\beta=\log_2(n)(1+\log_2(n))$ for Bitonic. The only hyperparameter optimized for Diffsurv is the choice of sorting network. Both sorting network were evaluated on the validation set but only the resulting models with higher validation c-index were used on the test set. This was repeated for each risk set size.

For the real-world datasets, the hazard function $h_\theta$ and $f_\theta$ for Diffsurv is Multi-layer Perceptron network. Hyperparameters: number of hidden layers, size of hidden layers, dropout rate, batch size and learning rate are determined by a small hyperparameter sweep of 100 trails on random 80:20 splits for each dataset and method. Hyperparameter optimization for Diffsurv also included sorting network and steepness. 

Full hyperparameter ranges and resulting best models are provided along with further implementation details at \url{https://github.com/andre-vauvelle/diffsurv-ea}.

During evaluation, the sorting network is not used since we only need to evaluate the ranks of the trained risk scores. Similarly, case-control sampling is not used. We measure the ranking performance of the models using the concordance index \citet{harrell_evaluating_1982}. 

\subsection{Possible Permutation Matrix}

\begin{figure}
    \centering
    \includegraphics[width=0.8\linewidth]{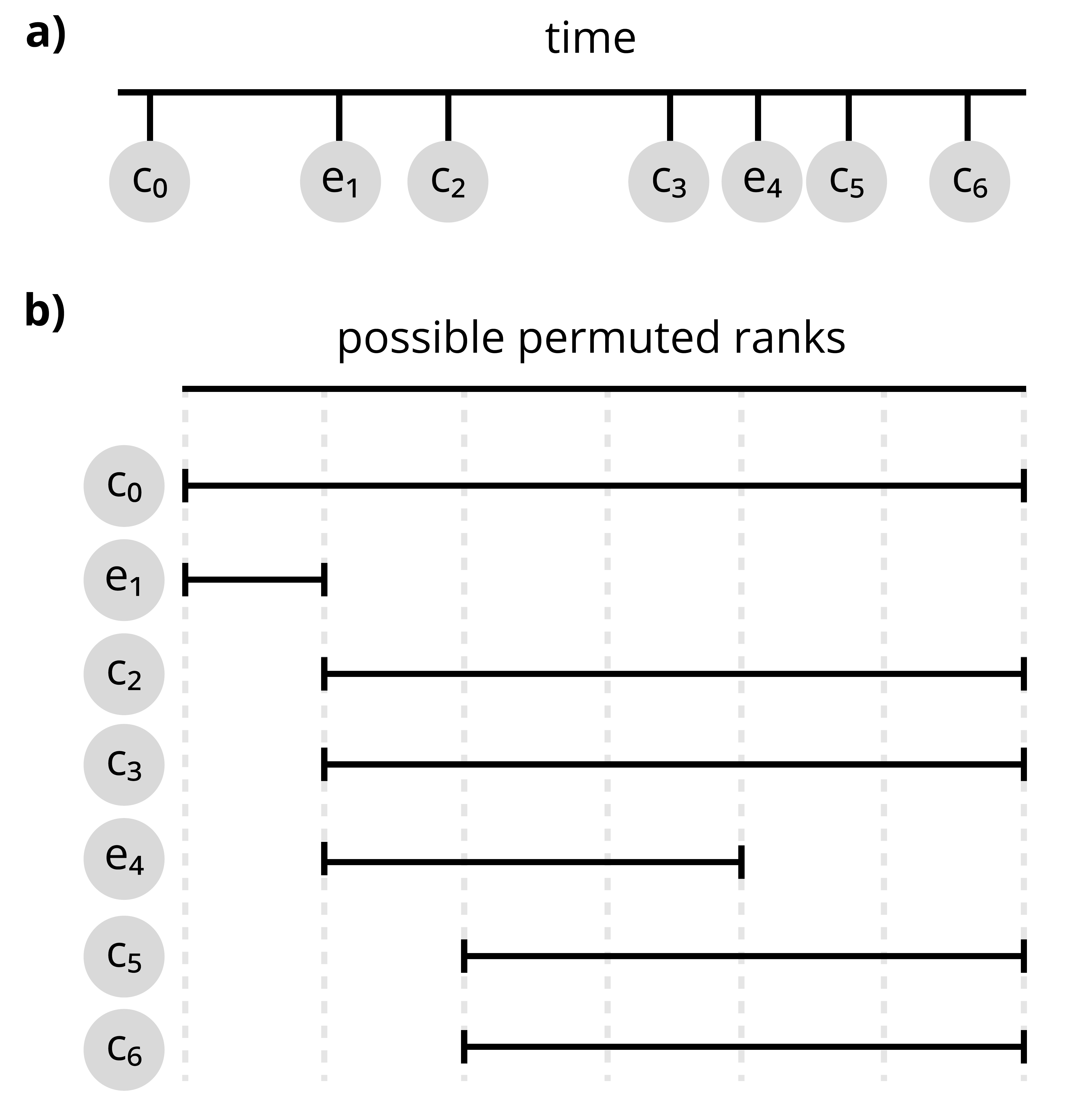}
    \caption{Possible permutations for an example case with two events ($e_0$ and $e_4$) and multiple censored samples ($c_0,c_2,c_3,c_5,c_6$).}
    \label{fig:possible-permutations}
\end{figure}

\begin{figure}
\centering
\subfloat[Possible Permutation Graph, $\gG_p$]{\includegraphics[width=10cm]{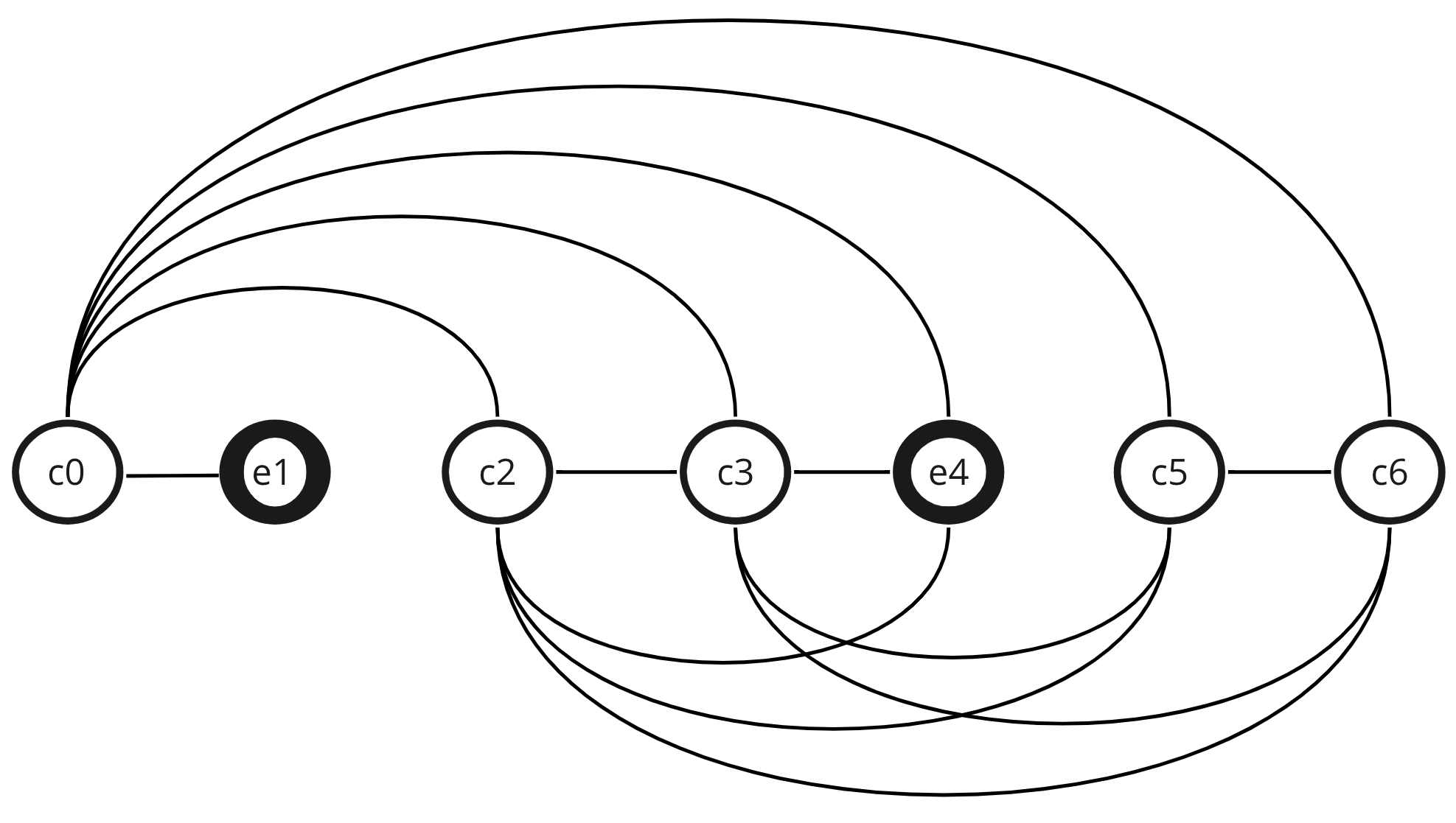}} 
\qquad 
\subfloat[Order Graph, $\gG_o$]{\includegraphics[width=10cm]{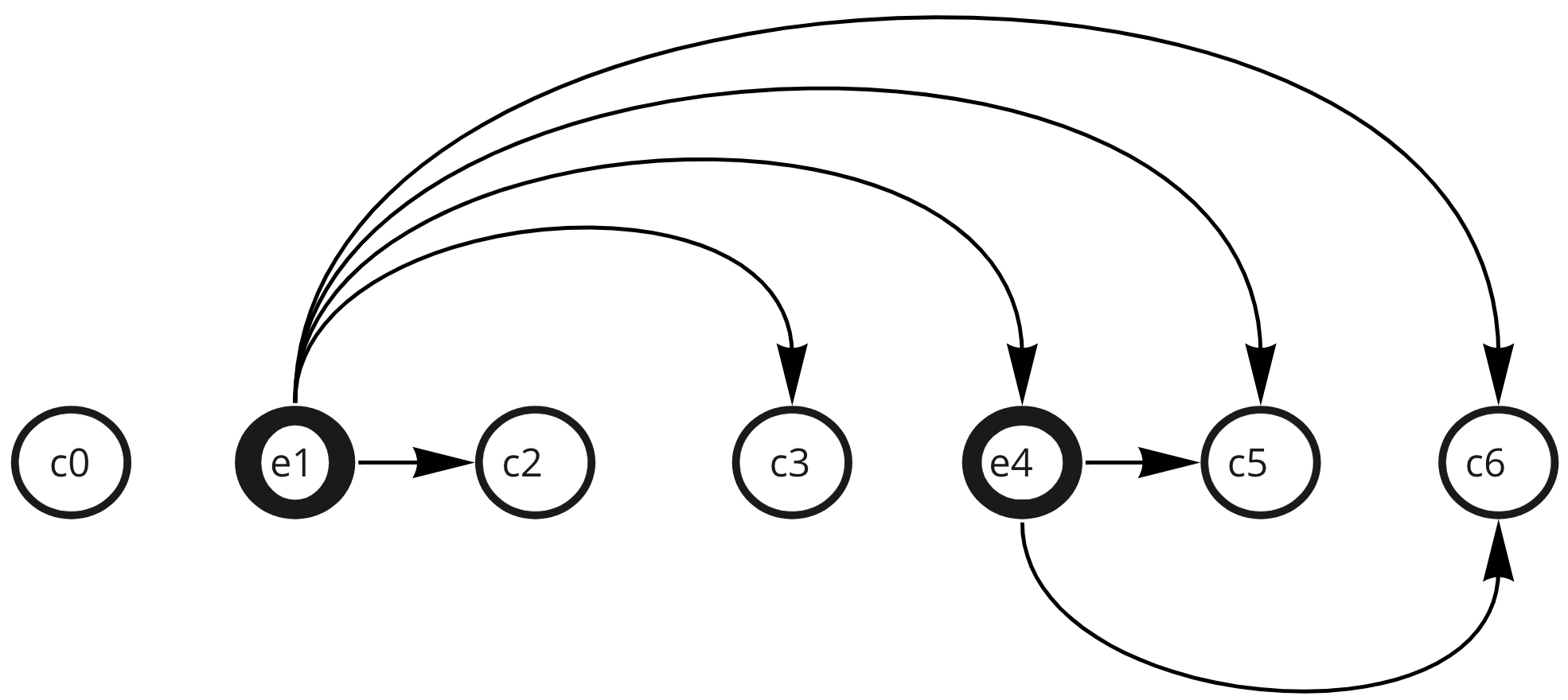}} 
\caption{Graphs representation of possible permutations and order.}
\label{fig:graphs}
\end{figure}

In order to account for censoring, we propose utilizing a possible permutation matrix $Q_p$. We provide a visualization for example risk set size 7 in Figure~\ref{fig:possible-permutations}, which corresponds with Equation~\ref{eq:possible-perm}. It is also possible to represent such connections as a graph. Further, this possible permutation graph $\gG_p$, is the complement of the order graph $\gG_o$ typically used for C-index. Both graphs are visualised in Figure~\ref{fig:graphs}

\begin{equation}
    \mQ_p = 
    \begin{pmatrix}
        1 & 1 & 1 & 1 & 1 & 1 & 1 \\
        1 & 1 & 0 & 0 & 0 & 0 & 0 \\
        0 & 1 & 1 & 1 & 1 & 1 & 1  \\
        0 & 1 & 1 & 1 & 1 & 1 & 1  \\
        0 & 1 & 1 & 1 & 1 & 0 & 0  \\
        0 & 0 & 1 & 1 & 1 & 1 & 1  \\
        0 & 0 & 1 & 1 & 1 & 1 & 1  
    \end{pmatrix}
    \label{eq:possible-perm}
\end{equation}

\subsection{Real World Datasets}
\label{apd:realworld}

\begin{itemize}
\item \textbf{FLCHAIN dataset:} A dataset containing information on patients with monoclonal gammopathy of undetermined significance (MGUS), focusing on serum free light chain (FLC) levels to study their prognostic significance in predicting disease progression.

\item \textbf{NWTS dataset:} A dataset from a series of clinical trials on the treatment and outcomes of children with Wilms' tumor, a type of kidney cancer, aiming to improve understanding of tumor biology and optimize treatment strategies.
\item \textbf{SUPPORT dataset:} A dataset from a multicenter study investigating the prognosis and treatment preferences of seriously ill hospitalized adults, with the goal of improving end-of-life care and informing decision-making processes.
\item \textbf{METABRIC dataset:} A dataset comprising genomic and clinical data on breast cancer patients, focused on uncovering novel molecular subtypes for more precise prognostication and personalized treatment strategies.
\end{itemize}

\subsection{SurvSVHN Dataset}
\begin{figure}
    \centering
    \includegraphics[width=\linewidth]{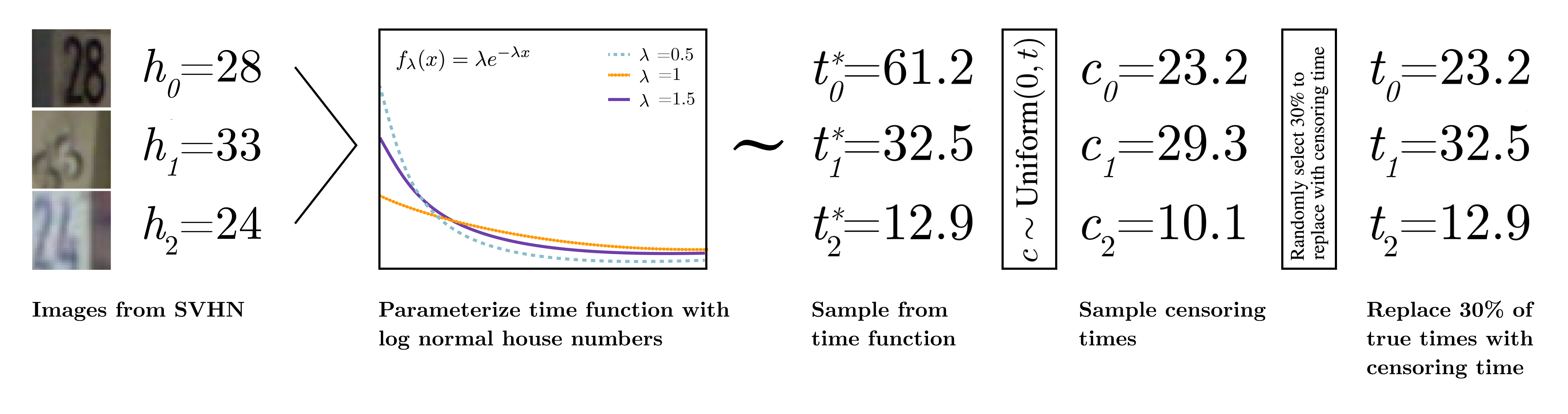}
    \caption{Visual abstract of the survSVHN dataset.}
    \label{fig:svnh-abstract}
\end{figure}





\end{document}